\documentclass[conference]{IEEEtran}
\usepackage[letterpaper, left=0.75in, right=0.75in, bottom=0.75in, top=0.75in]{geometry}

\usepackage{amsmath}
\usepackage{amssymb}
\usepackage{xparse}
\usepackage{xspace}
\usepackage{url}
\usepackage{graphicx}
\usepackage{subfig}
\usepackage{mathtools}
\usepackage{booktabs}

\usepackage{mathtools}
\usepackage{float}
\usepackage{mciteplus}
\usepackage{hyperref}

\usepackage{ragged2e}
\usepackage{colortbl}
\usepackage{xcolor}

\def\andothers{et al.\,}
\def\figname{Fig.\,}
\def\secname{Section\,}
\def\tabname{Table\,}
\def\mocap{Motion Capturing System\xspace}
\IEEEoverridecommandlockouts                                

\title{\LARGE \bf Simitate: A Hybrid Imitation Learning Benchmark}

\graphicspath{{images/}}

\author{Raphael Memmesheimer, Ivanna Mykhalchyshyna, Viktor Seib, Dietrich Paulus\\
Active Vision Group, Institute for Computational Visualistics,
        University of Koblenz-Landau, Germany\\
       {\tt\small \{raphael, ivannamyckhal, vseib, paulus\}@uni-koblenz.de}\\
}

\begin{document}

\twocolumn[{
\renewcommand\twocolumn[1][]{#1}
\vspace{2em}
\maketitle
\begin{center}
    \centering
    \vspace{-1em}
    \includegraphics[width=.8\linewidth]{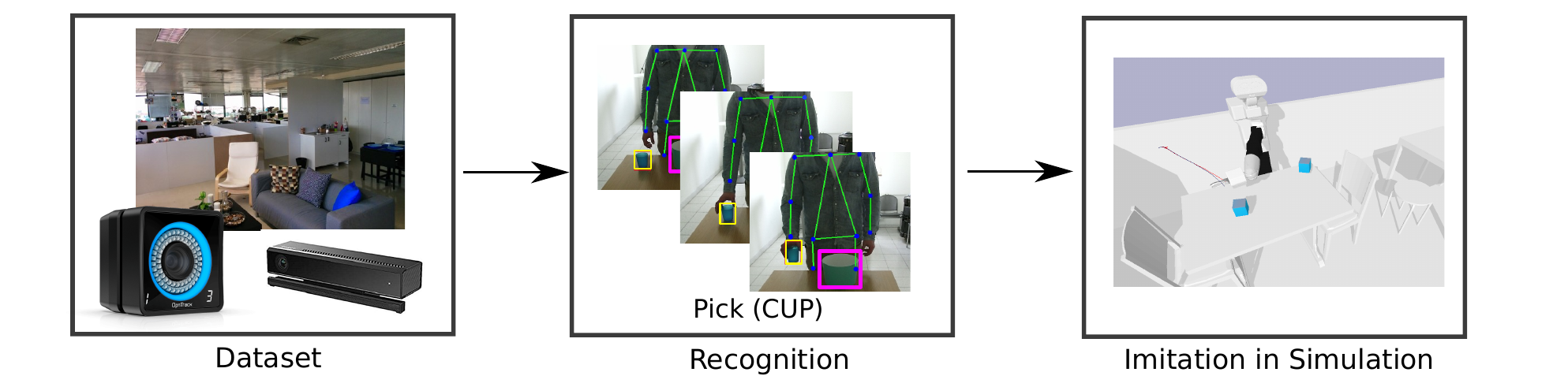}
    \captionof{figure}{Overview: This figure gives an overview of our benchmarking model. We provide a dataset
        recorded with a RGB-D camera and a motion capturing system. The sequences of the dataset are supposed
        to be interpreted by approaches for imitation learning which then have to execute the imitation
        in a simulated environment based on initial object
    positions of the provided ground truth. After the performance in simulation, results are automatically evaluated.}
    \label{fig:overview}
\end{center}%
}]


\IEEEpeerreviewmaketitle
\thispagestyle{empty} 
\pagestyle{empty}

\thispagestyle{empty}
\pagestyle{empty}

\begin{abstract}
    We present Simitate --- a hybrid benchmarking suite targeting the evaluation
    of approaches for imitation learning. A dataset containing 1938
    sequences where humans perform daily activities in a realistic environment is
    presented.  The dataset is strongly coupled with an integration into a    simulator.
    RGB and depth streams with a resolution of 960$\mathbb{\times}$540 at 30Hz and accurate ground truth poses 
    for the demonstrator's hand, as well as the object in 6 DOF at 120Hz are provided.
    Along with our dataset we provide the 3D model of the used environment, 
    labeled object images and pre-trained models. 
    A benchmarking suite that aims at fostering comparability and reproducibility supports the development of imitation learning approaches.
    Further, we propose and integrate evaluation metrics on assessing the quality of effect and 
    trajectory of the imitation performed in simulation. Simitate is available on our project website: \url{https://agas.uni-koblenz.de/data/simitate/}.

    

\end{abstract}

\section{Introduction}
\label{sec:introduction}

The application of robots in domestic environments is foreseeable and we believe
that with the future spread of robots the demand for custom service robot tasks
and therefore expert programmers will increase dramatically. We therefore publish
a dataset that fosters imitation learning approaches just by visual observation
of humans interacting with their environment. This supports the demonstrator when
interacting in a natural way with its environment (let it be objects or humans).
This idea stands in high contrast to current approaches that pull demonstrators
out of their natural interaction by putting sensor suites or use kinesthetic teaching of
robots. The Programming by Demonstration paradigm is most famous for various applications
in industrial repetitive task programming. 

Motivated by the increasing success of deep neural networks that recently opened
up possibilities for reasonable accurate object recognition \cite{he2016deep, krizhevsky2012imagenet, szegedy2015going},
detection \cite{redmon2016you, liu2016ssd}, semantic segmentation \cite{he2017mask, badrinarayanan2015segnet, DeepMask}
and human pose estimations \cite{cao2017realtime, simon2017hand, wei2016cpm}, we believe in advancing these fundamental
approaches to actual scene understanding and even replication with a mobile domestic
service robot. 

As of now, imitation learning approaches are often empirically evaluated and show qualitative
results that are commonly demonstrated on a small set of actions. There is no
common dataset available that allows for comparison of approaches on a standardized dataset
as is the case for many other topics like image classification \cite{krizhevsky2012imagenet}, 
object detection \cite{lin2014microsoft}, object tracking \cite{bernardin2008evaluating} or 
position estimation \cite{geiger2012we, sturm2012benchmark}.
This might be caused by the complexity of the evaluation process for the imitation learning
task. We try to tackle this problem in this work and aim at providing a dataset and benchmarking
combination that supports the development and evaluation of imitation learning tasks.
To the best of our knowledge, there are no commonly used metrics for evaluation 
of imitation learning tasks. The importance of such a metric has been highlighted already in 2009
\cite{argall2009survey} and again in 2018 \cite{osa2018algorithmic}.
We found robotic imitation learning approaches that use custom
collected data for experiments, but this data has not been published for general access.  
This makes reproduction and comparison harder or even impossible.

In high contrast to other currently available datasets we do not only focus on the
recognition of actions, but also on a deeper understanding of the interaction
between humans and objects. Even though we also recorded
ground truth positions of the demonstrator's hand and the interacting
objects, the goal of the benchmark is to advance in markerless visual imitation 
learning approaches.

Simitate will be applicable for approaches in different fields like imitation learning 
through reinforcement learning \cite{duan2017one}, genetic programming 
\cite{gangwani2018policy} or generative adversarial networks \cite{gail2015towards}.
Beside imitation learning, the dataset can be used for action recognition or object
tracking, but does not primarily target this fields. 



The main contributions of this paper are:
\begin{enumerate}
    \item a novel publicly available dataset containing different individuals performing daily activity
tasks
    \item a novel benchmarking component which enables researchers to compare their results
in a simulated environment
    \item metrics for evaluation based on the imitated trajectory and the resulting effect are proposed.
\end{enumerate}

The paper is structured as follows. First we introduce related work in terms
of datasets as well as approaches in this field in \secname \ref{sec:related_work}. 
In \secname \ref{sec:dataset}, we then describe the dataset we provide as well as the setup and
sensors we used during recording. 
We extend the dataset by providing a simulative benchmark and suggest metrics
in \secname \ref{sec:benchmark}.
Finally, we conclude this paper in \secname \ref{sec:conclusion}.

\section{Related Work}
\label{sec:related_work}
Most approaches use
custom datasets and methods for evaluation, making direct comparisons vague.
Ross \andothers \cite{Ross2010} presented a supervised approach for imitation learning by dataset aggregation, called DAgger.
Expert policies which gather a dataset of trajectories are used to train a second policy that aims at mimicking the trajectories well. 
Afterwards, more policies from expert demonstrations are used again to mimic the demonstrations but now the trained policies are added to the dataset.
The next policy is then defined as the policy that best mimics the expert on the whole dataset.
Laskey \andothers \cite{Laskey2017} proposed an off-policy approach which injects noise
into the demonstrator's policy. By this the demonstrator is forced to correct the injected noise and a
recovery behaviour from errors can be trained. In comparison to DAgger \cite{Ross2010} they claim the 
approach to be faster and more robust. The data from the physical experiments
on a real robot is not available.  
Ho \andothers \cite{ho2016generative} presented an approach for 
extracting policies directly from data by a model-free imitation learning algorithm. 
Their approach has been proven to show same results as inverse reinforcement learning problems.
One shot imitation learning approaches \cite{DBLP:conf/rss/YuFDXZAL18, DBLP:journals/corr/abs-1810-05017, DBLP:conf/nips/DuanASHSSAZ17} have recently gained popularity.
Further virtual reality approaches have been used for learning new activities by 
demonstration \cite{bates2017line,amaro2014bootstrapping}.
A promising crowd sourcing approach of human-robot interactions was proposed by Mizuchi \andothers \cite{mizuchi2017cloud}. This 
potentially could enable learning robot activities by  demonstrations through virtual reality.
All virtual reality approaches lack the direct transferability to real world
robots because of the usage of simulated sensor data. We try to tackle this bottleneck
in this paper.
Comparable datasets mostly target action recognition or classification approaches.
Weinzaepfel \andothers \cite{weinzaepfel2016human} presented DALY, a dataset containing
ten daily activity classes found in 500 youtube videos with a total duration of 31 hours.
Pirsiavash \andothers \cite{pirsiavash2012detecting} created a first person dataset containing images
from people fulfilling daily activity tasks.  Most datasets focus on action 
recognition, a comprehensive survey is given by Zhang \andothers \cite{zhang2016rgb}. 
Many published datasets focusing on imitation learning target autonomous driving
\cite{Codevilla2018} \cite{zhang2016query}.
Gupta \andothers presented a dataset \cite{gupta2015visual} based on 
a subset of the COCO \cite{lin2014microsoft} dataset targeting semantic role labeling by verbs describing
people interacting with objects.
The dataset that comes closest to our proposed dataset is the CAD-120 by Koppula 
\andothers \cite{koppula2013learning} which contains 120 different RGB-D camera 
sequences where four individuals perform activities like making cereal, microwaving 
food and more.  In addition, the dataset contains skeleton data provided by a skeleton 
tracker and manually annotated object tracks.

Benchmarking nowadays enables quantitative evaluation in many research topics
like autonomous driving \cite{geiger2012we}, RGB-D SLAM systems \cite{sturm2012benchmark},
object tracking \cite{bernardin2008evaluating, wu2013online, milan2016mot16}.
Those benchmarks build a comfortable environment for evaluation as most commonly
standard formats, evaluation metrics and scripts are specified for result comparison. 
Most of them even collect produced evaluation results online \cite{geiger2012we, milan2016mot16}
in a leader board.
Some of the later benchmarks also integrate the replication by actual robotic systems
i.e. for grasping \cite{leitner2017acrv}.
Virtual reality environments have been previously used \cite{zhang2017deep,Villani2018}
for evaluation of human robot interfaces.
In form of competitions like RoboCup@Home \cite{wisspeintner2009robocup} 
robotic systems are benchmarked in domestic environments, however, due to the biannual 
changes of the rules and not fully objective opinions of referees the comparison should be seen critical. 
Further, the focus is set on a time constrained one shot evaluation in most tasks.
In contrast, the European Robotics League \cite{lima2016rockin}
puts a focus on benchmarking and uses explicit metrics. However, long 
term benchmarking and the limited amount of participating teams still makes long term
comparability hard.
Some metrics have been proposed mainly for the correspondence problem of imitation
learning tasks \cite{Alissandrakis2007}. A promising approach is to measure the
effect based on \cite{Alissandrakis2006} where demonstrated and imitated effects are
compared by their displacements in relation to other objects. Most common for
the evaluation of imitation learning tasks are qualitative observations \cite{Ross2010,Laskey2017}.
This is a major deficit in comparison to other well established fields.



\section{Dataset}
\label{sec:dataset}

\begin{figure*}[t] \centering$
  \vspace{0.06in}
  \begin{array}{cccc}
      \includegraphics[height=3.15cm]{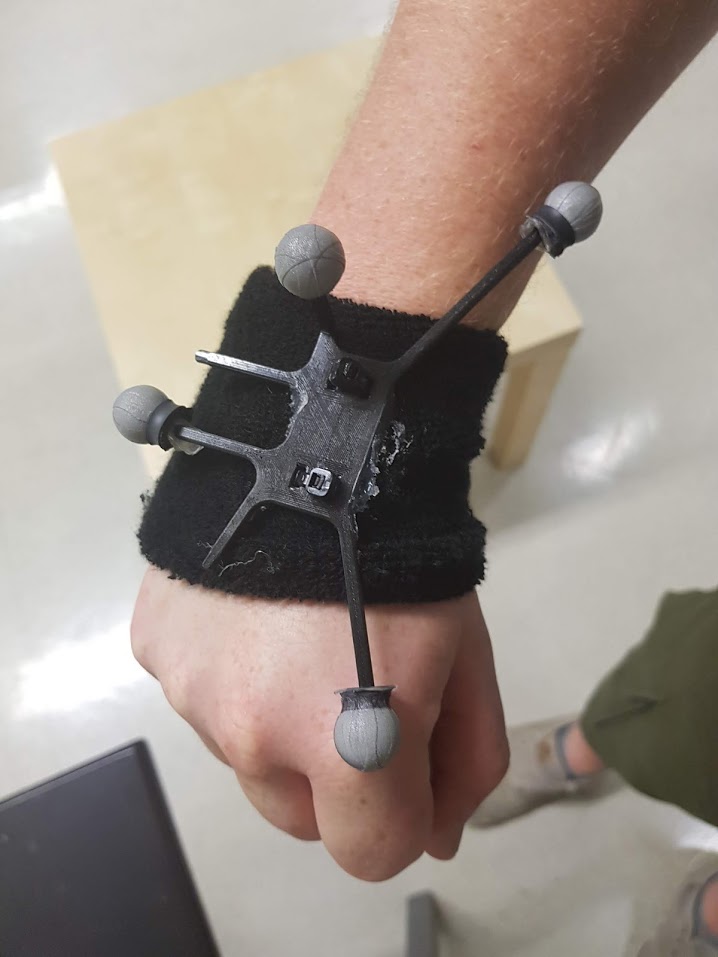} &
      \includegraphics[height=3.15cm]{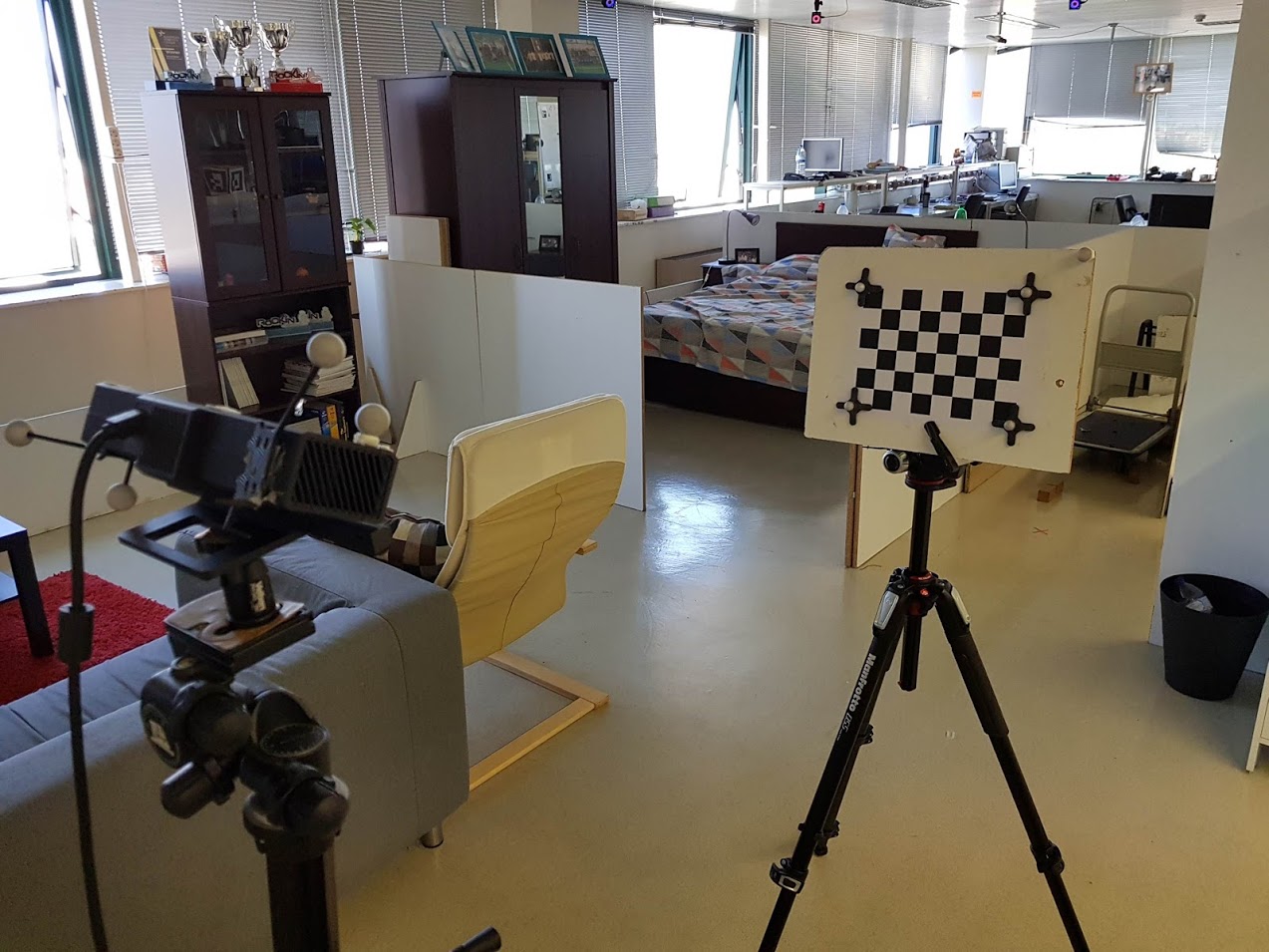} &
      \includegraphics[height=3.15cm]{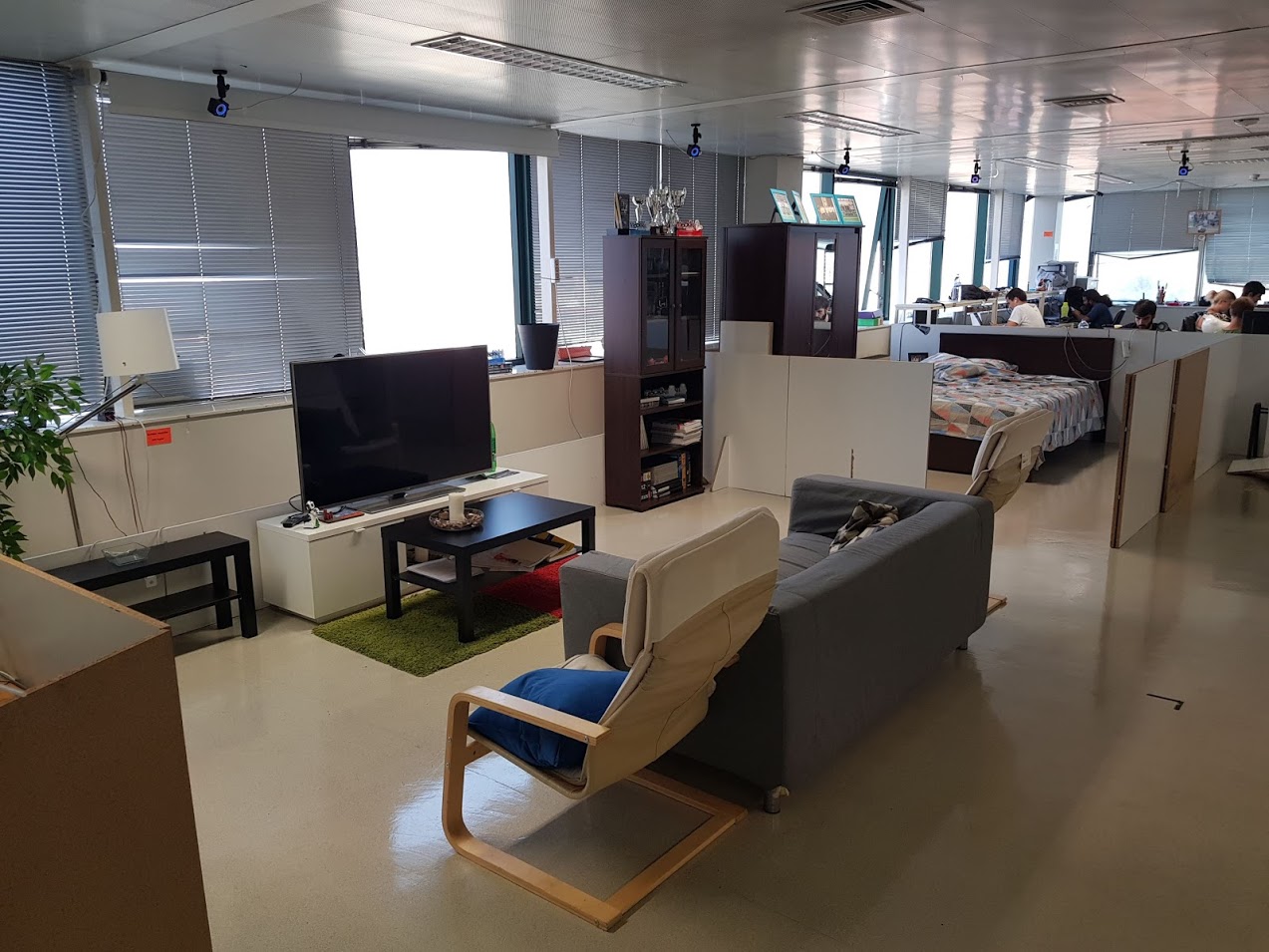} &
      \includegraphics[height=3.15cm]{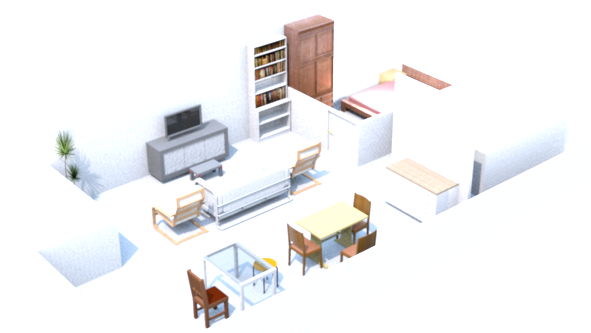}\\
      (a) & (b) & (c) & (d) 
  \end{array}$
  \caption{Dataset setup. Reflective markers are attached on the humans hand $(a)$ and the interacting objects $(b)$. In $(c)$ an exemplary demonstration is shown and $(d)$ shows a rendered view of the apartment used, which also will be used in the simulated environment of the benchmark.}
  \label{fig:setup}
\end{figure*}

In this section we describe the setup for the dataset acquisition, including the applied testbed and motion capturing system setup.
Further, the dataset's resulting sequences are introduced.

\subsection{Setup}

To record the dataset we used a Kinect 2 RGB-D camera mounted on a tripod.
Data was acquired in an exemplar 
apartment modelling common real world apartments, including different furniture 
and rooms.
12 "OptiTrack PRIME 13" cameras were mounted on the ceiling. In total an area of 50$m^2$ is
covered by the system. The optical center of the RGB-D camera is calibrated 
against the motion capturing system. Rigid
body markers are attached to all relevant interacting objects and the human demonstrator.
The demonstrator is completely visible during recording, except when he is occluded
by objects or furniture that he is interacting with. The individual sequences were recorded at a number of
different locations in the apartment. For inspection purposes we also recorded
a camera stream giving an overview of the apartment.

\subsection{Calibration}

\begin{figure}[t]
\centering
  \includegraphics[width=.8\linewidth]{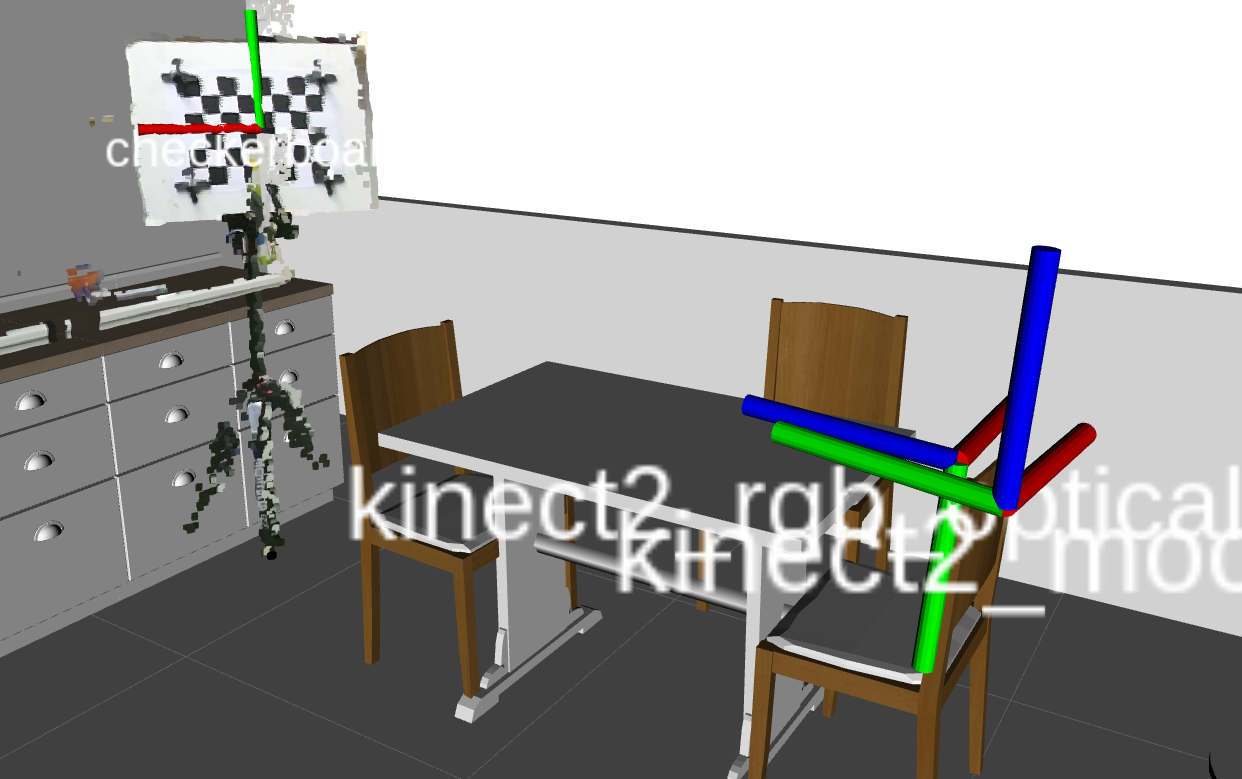}
  \caption{A visualization of the resulting calibration. The checkerboard center is correctly aligned to its corresponding \mocap marker.}
  \label{fig:reprojection}
\end{figure}

The motion capturing system has been carefully calibrated before recording
the sequences using OptiTrack Motive motion capturing software with a CW-500
marker. A common origin has been estimated using a CW-200 marker in a fixed point 
of the apartment. For the motion capturing calibration we achieved the following 
results. The mean overall wand error was $0.136mm$. For re-projection we got a mean 3D 
error of $0.523mm$, and a mean 2D error of $0.099$ pixels. The worst mean re-projection 
3D error was at $0.642mm$ and the worst mean 2D error was at $0.143$ pixels.
The RGB-D camera has been calibrated intrinsically and extrinsically. 
For calibration of the RGB-D camera against the \mocap
we followed Sturm \andothers \cite{sturm2012benchmark} ideas. Reflective markers were 
attached at the corners of a checkerboard pattern. The centroid of the
checkerboard was estimated using the \mocap and the central checkerboard pixel 
for corresponding image coordinates. It was ensured that the printed pattern was
completely planar. We estimated the reflective marker height using the CW-200 
marker and updated the centroid to be on the same planar surface as the
printed pattern. \figname \ref{fig:reprojection} shows reprojected marker of the checkerboard center.
The transformation between the centroid of the RGB-D camera's rigid body and 
optical center of the RGB-D camera are then estimated \cite{zhang2000flexible}.
In order not to interfere with the calibration result by motion we mounted the RGB-D camera 
and the checkerboard on tripods. The inverse transformation is used between the 
\mocap pose of the RGB-D camera and its optical center.
A precise calibration is especially important for the alignment of real world data and later 
imitation in simulation. Too high residuals will lead to an inaccurate alignment between simulation and read world observation and can affect the imitation performance. 

\subsection{Testbed}
\label{sec:testbed}




The testbed ISRoboNet@Home\footnote{\url{http://welcome.isr.tecnico.ulisboa.pt/isrobonet/}} has been set up for the European Robotics League to support 
the benchmarking of service robots. It aims at imitating a domestic environment 
separated in different rooms, including
standard furniture and objects.
The \mocap system described above is integrated in the testbed and allows to record ground truth data of interacting humans, robots, as well as objects.
Besides the installed \mocap this testbed has the following benefits: its initial state can be recovered, it is similar to real apartments and it is open for use by
research groups. This benefits also allow everyone to extend the set of recorded sequences.



\subsection{Human-Object interactions}

\begin{figure}\centering
  \includegraphics[width=.8\linewidth]{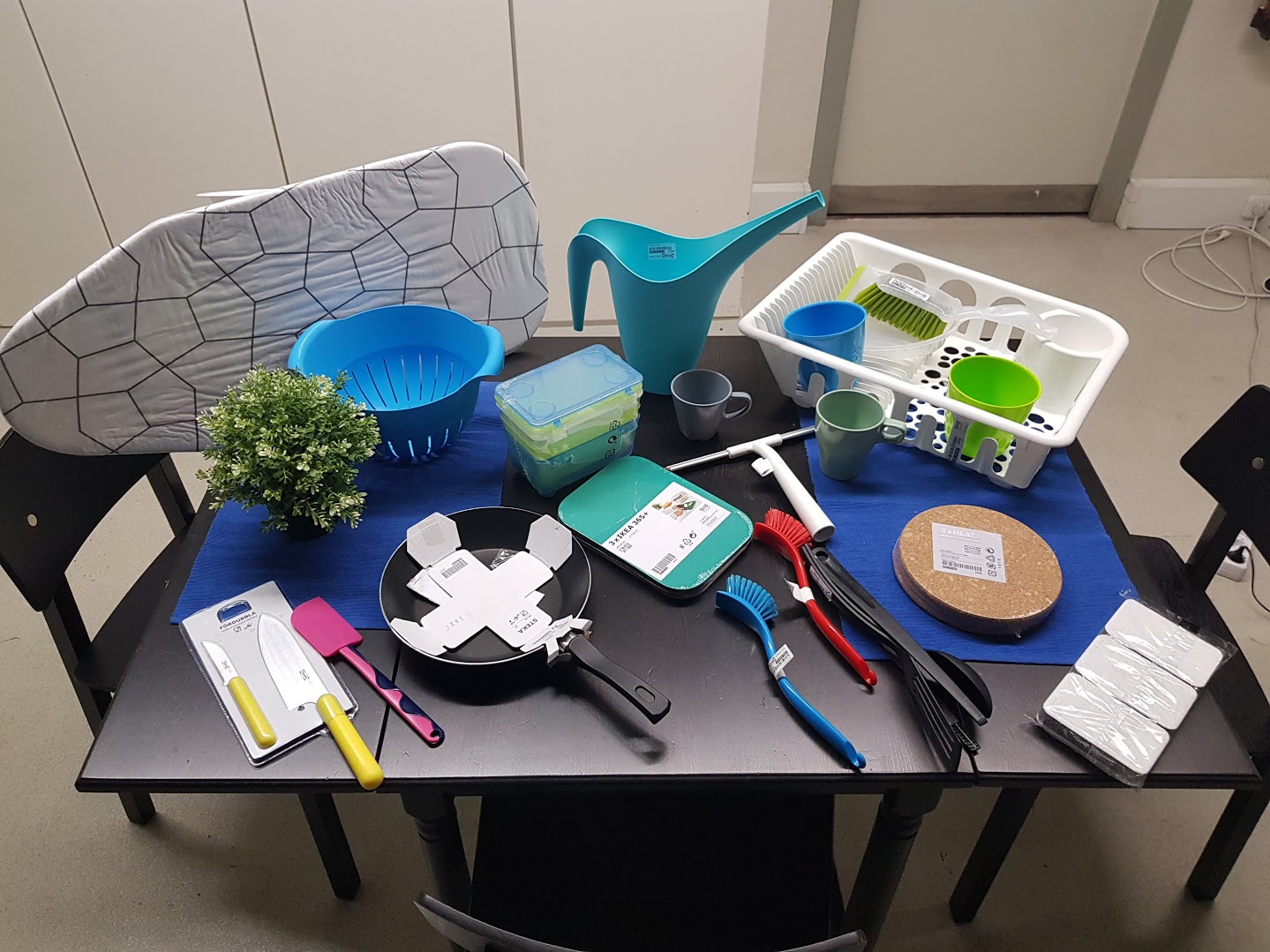}
  \caption{Objects used for the dataset}
  \label{fig:objects}
\end{figure}

We used common affordable home accessories that we got from a worldwide serving
furniture retailer. The used objects are depicted in \figname \ref{fig:objects}. 
For reproduction we also provide a list of objects, including 
their labeled training images and pre-trained models for two widely
spread recent approaches \cite{redmon2016you, he2017mask}. The images have been labeled
with support of a recent guided image segmentation approach \cite{Man+18}.
The provided data allows to easily reproduce the results and deminished the hurdles to develop approaches for this benchmark. We tried to get 
colorful objects too, as the focus of the presented benchmark should not be on object 
recognition, but on the imitation learning aspect. 


We mount rigid body markers at the back of the right hand of the demonstrator.
An exemplary setup for the human is shown in \figname \ref{fig:setup} $(a)$. 
We ensured that human pose estimates using a recent key-point detector are not 
interfered with by the marker setup.
We provide human body keypoints extracted with OpenPose\cite{wei2016cpm} and 
projected using the depth channel into world coordinates as well.


\subsection{Sequences}

We recorded sequences for multiple purposes. First, we want to ensure that different
categories of imitation learning can use this dataset. Therefore, we recorded sequences
that aim at the interpretation of the demonstrations on a symbolic as well as on
a trajectory level. 

Sequences on a trajectory level are further divided into cloning tasks, where the
human performs a movement and the goal is to mimic the movement. More challenging
sequences contain object interactions.
All sequences are performed by different individuals. We provide sequences
that cover not only local demonstrations, but also movements between different places
in the apartment of \figname \ref{fig:setup} $(d)$. 
For tasks like opening a door, we ensured to handle multiple doors of the apartment.
We divide the sequences based on their level of difficulty. \textit{Basic Motion} sequences
contain drawn figures with the right hand. Its intention are to clone the observed movement. They 
also serve as testing sequences for the hand position estimation.
\textit{Motion} sequences contain activities like reaching for an object with the hand, picking, placing, moving or pushing it. 
More complex activities contain tasks that are categorized as \textit{Complex} sequences. 
\textit{Sequential} scenes contain multiple basic motions in various random combinations 
over a longer period of time.
The complete list of sequences is given in \tabname \ref{tab:sequence_overview}.

\begin{figure*}[t] \centering$
    \vspace{0.06in}
    \begin{array}{ccc}
        \includegraphics[width=.3\linewidth]{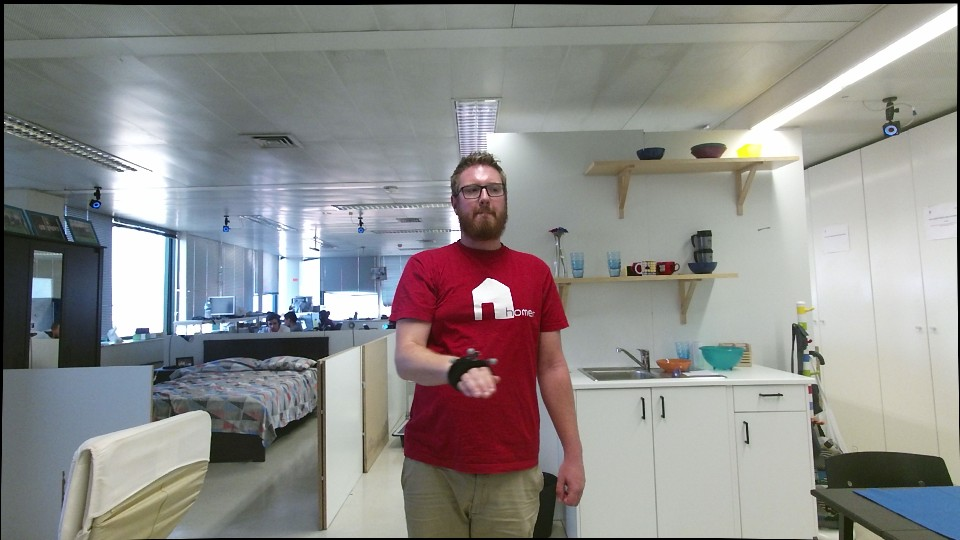} &
        \includegraphics[width=.3\linewidth]{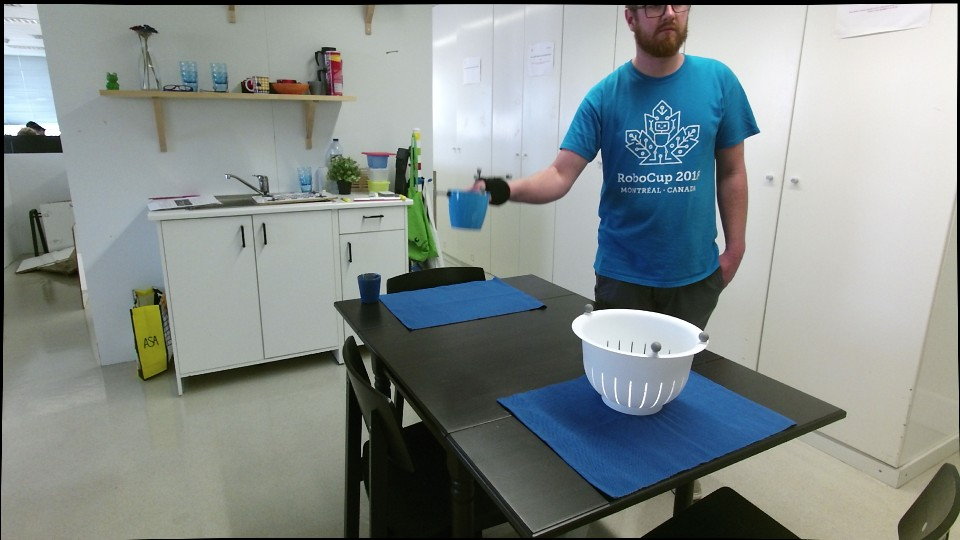} &
        \includegraphics[width=.3\linewidth]{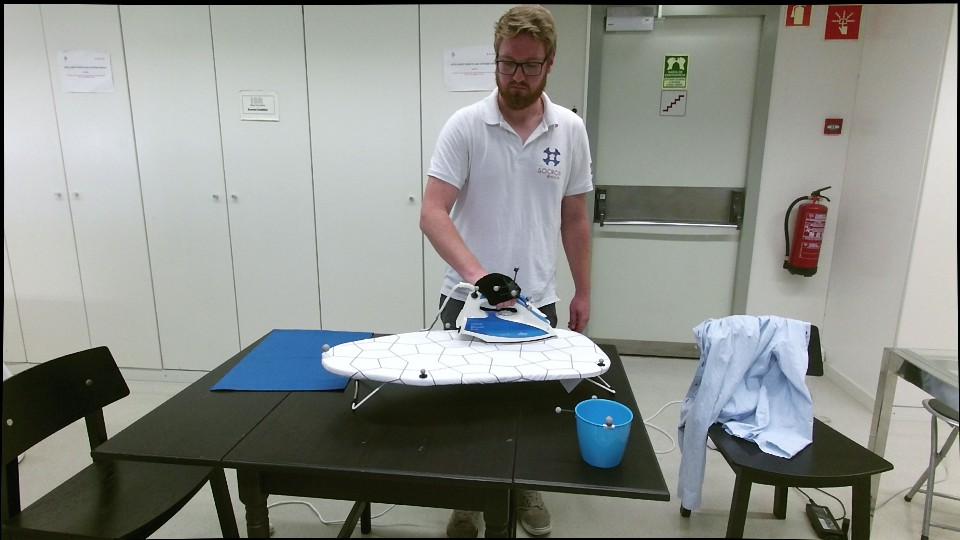} \\
        \includegraphics[width=.3\linewidth]{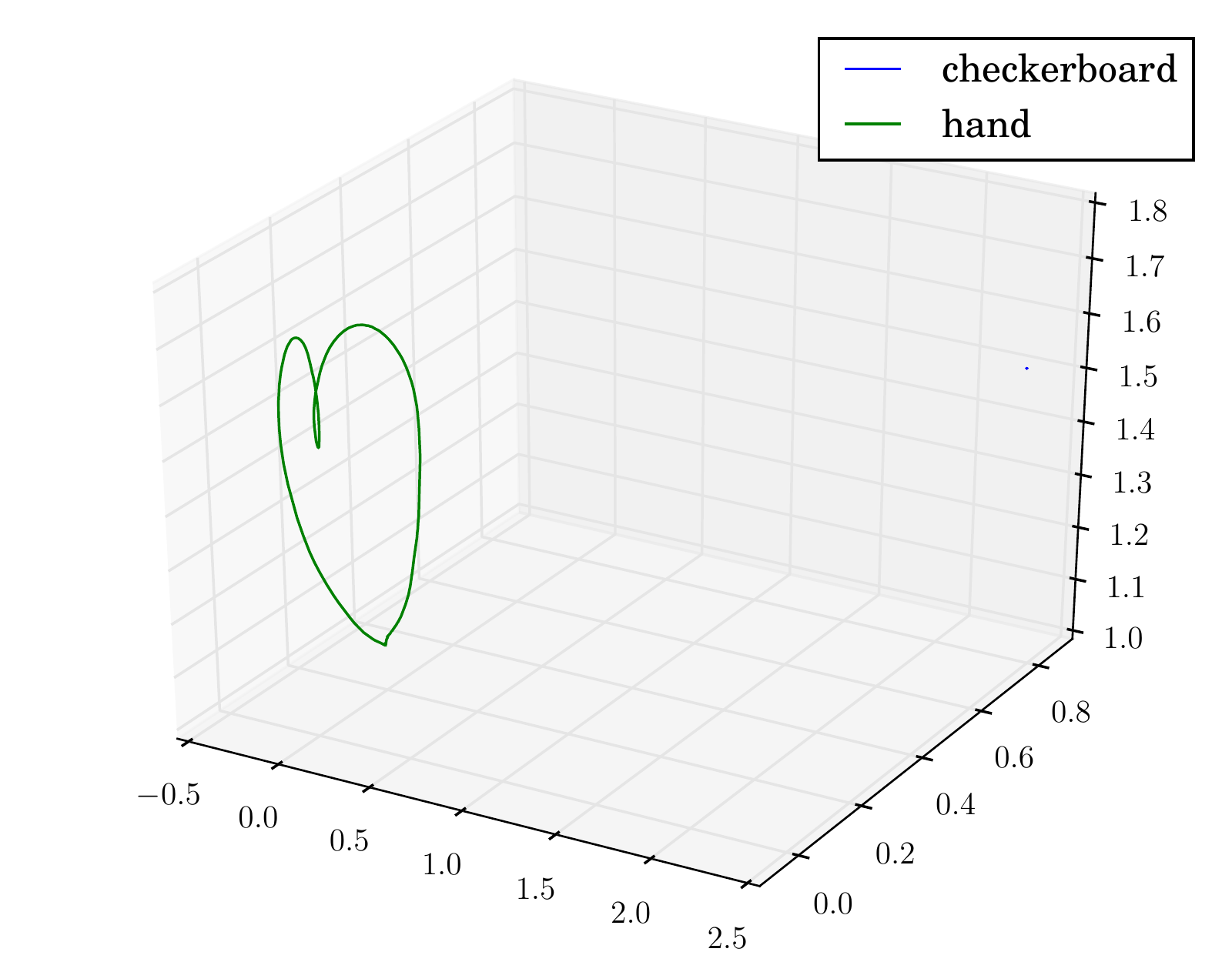} &
        \includegraphics[width=.3\linewidth]{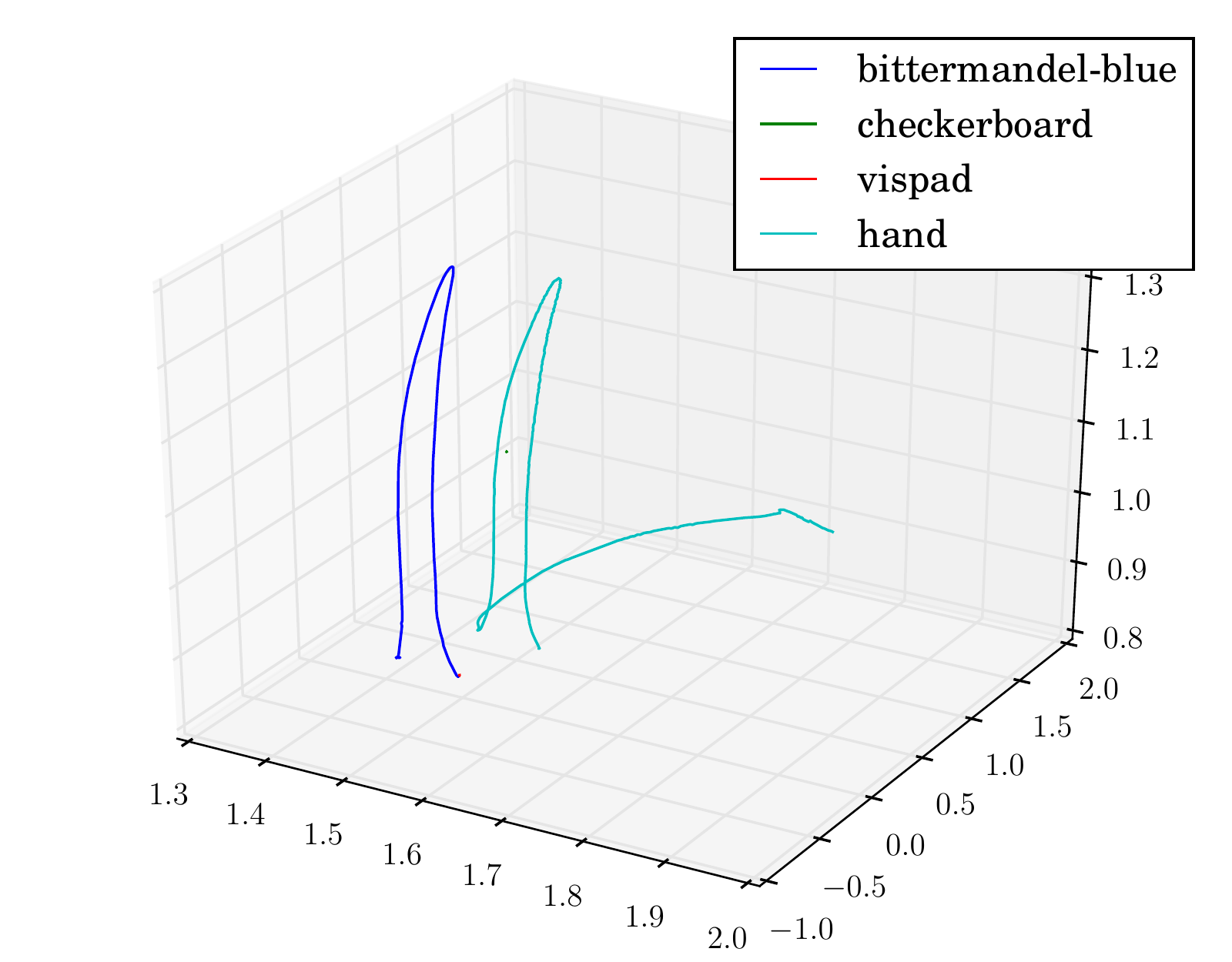} &
         \includegraphics[width=.3\linewidth]{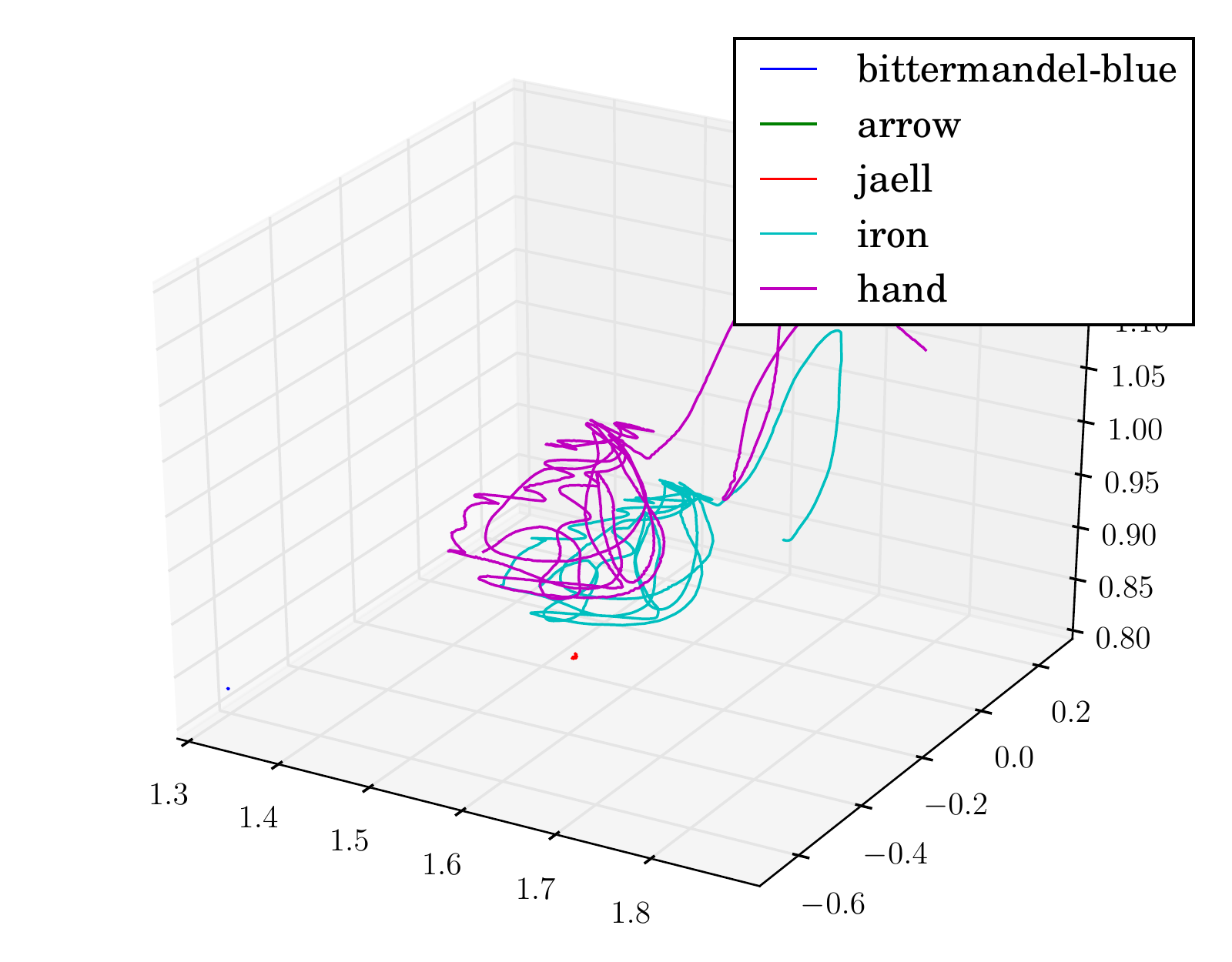} \\
        (a) & (b) & (c) 
    \end{array}
    $
    \caption{Example sequences image on top and plotted trajectories at the bottom for $(a)$ a basic motions heart sequence, $(b)$ a motion sequence for reaching, $(c)$ a complex sequence for ironing.}
    \label{fig:trajectories}
\end{figure*}

\begin{table}[]
  \vspace{0.06in}
    \centering
    \caption{Sequence overview}
    \newcommand{\bs}[1]{\textbf{#1}}
    \definecolor{SeqColor}{rgb}{0.9,0.9,0.9}
    \begin{tabular}{%
            >{\quad}
            p{0.25\linewidth}
            >{\RaggedLeft\arraybackslash}p{0.08\linewidth}
            >{\RaggedLeft\arraybackslash}p{0.2\linewidth}
            >{\RaggedLeft\arraybackslash}p{0.2\linewidth}
            }
        \toprule
         & \textbf{\# Seq} & \textbf{Avg. Length in s} & \textbf{Total Length in m} \\
        \midrule
            \rowcolor{SeqColor}
            \hspace{-1.5em} Basic Motions& & & \\
                Circle         & 104  & {6.83} & {11.58} \\
                Rectangle      & 105  & {6.84} & {11.97} \\
                Heart          & 85  & {6.85} & {9.70} \\
                Triangle       & 85  & {6.85} & {9.70} \\
                Zickzack       & 85  & {6.83} & {9.68} \\
            \rowcolor{SeqColor}
            \hspace{-1.5em} Motion& & & \\
                Reach         & 79      & 7.97 & {10.49} \\
                Move          & 79      & 7.96 & 10.48  \\
                Push          & 30      & 9.40 & 4.70  \\
                Pick          & 79      &     7.97  & 10.49  \\
                Place         & 79      &      7.96 &  10.49 \\
                Pour          & 224      &      8.25  &  30.83 \\
                Stack        &  63       &  14.63   & 15.36  \\
                Wipe         &  31     &    29.06 & 15.01  \\
                Mix          &  33      &   14.36  &  7.90 \\
            \rowcolor{SeqColor}
            \hspace{-1.5em} Complex& & & \\
            Ironing        & 92   &        31.74   &  48.66  \\
            Clean          & 92   &         28.11  &  43.11  \\
            Throw          & 50   &         6.84    &  5.70  \\
            Cut          & 49   &         19.37    &  15.82  \\
            Open          & 40   &         9.37    &  6.24  \\
            Close          & 20   &         4.34    &  1.44  \\
            \rowcolor{SeqColor}
            \hspace{-1.5em} Sequential& & &  \\
                Rearrange    & 65   &   19.33     & 20.94  \\
                Pick and Place   & 409   &   14.21 &  96.91 \\
                Place into   & 60     &     10.67  & 10.67  \\
                Bring        & 82     &     21.02  & 28.73  \\
        \bottomrule
    \end{tabular}
    \label{tab:sequence_overview}.
 \end{table}

\section{Benchmark}
\label{sec:benchmark}
\def\rpe{RPE\,}
\def\ape{APE\,}

%
We propose a combined approach of real world observations and simulated environment
for benchmarking imitation learning approaches. The initial object locations 
and positions of the observing sensors are propagated into a carefully reconstructed simulation
of the testbed.
This approach has multiple benefits: First, this enables evaluation methods for
imitation learning and extends currently available datasets that focus
on action recognition. Further, it supports generalization as the imitated behavior
could be benchmarked with a wider variety of simulated robots and simplifies the 
transfer to real world robots. Furthermore, it 
enables generalization to verify the imitated behavior with a variety of objects
and locations.
Exemplary, we provide integration into two widely used simulations \cite{koenig2004design, coumans2019} 
in the robotics and machine learning community. The benchmark in combination 
with the provided dataset therefore allows the evaluation of action recognition 
and task imitation on semantic and trajectory level. As action recognition
is already addressed by many other datasets, we focus on the imitation aspect in
the benchmark description.

To reduce the complexity in application of this benchmarking approach and to foster the development of
imitation learning approaches we provide labeled training data for object segmentation and object detection
as well as pre-trained models for current state of the art approaches \cite{he2017mask, redmon2016you}.
The benchmark is supposed to be executed sequentially. First, the individual
sequences are played back. This sequence has to be analyzed by an approach
either on semantic or trajectory level. After the analysis, the task is reproduce the observed actions. Generalization is evaluated by replication of the same 
tasks using different initial setups, but common actions on previously unseen sequences.
In the observation step, sequences from the dataset will be analyzed and relevant information
for the recognized action, interacting objects and arm trajectories should be extracted. We provide a class that simplifies this for later evaluation.
The ground truth information from the sequence is used to initially setup  the 
virtual representation of the testbed in simulation. A simulated robot should then execute the
observed action. This allows evaluation of the achieved
effect and trajectory error measurements. 

\subsubsection{Effect}


Using the effect has been proposed by Allisanrakis \andothers \cite{Alissandrakis2007}. We
 integrate effect evaluation for relative and absolute effects after 
performance of the imitation. Evaluating the relative object pose
seems to be appropriate when objects are placed very close to each other. In this case we
can measure the relative pose error $\rpe$ between the final object pose $p_e$ and the
relative ground truth poses between the object and the $j$th of $n$ surrounding objects $p_{g,j}$ like:

\begin{equation}
    {\rpe}:= \sum_{j=1}^{n} (p_{e} \ominus p_{g,j})^{2},
\end{equation}
where $\ominus$ is the inverse motion compensation operator \cite{kummerle2009measuring} that can be imagined as the relative 3D transformation between two poses. 
This metric is inspired
by suggestions for the accuracy of SLAM systems \cite{kummerle2009measuring, sturm2012benchmark}.
The success of the imitation is evaluated based on post conditions
that are modeled by the end state of the ground truth. 
In other cases it will be more relevant to aim for an effect in the humans coordinate 
frame. For that we use the absolute pose error:

\begin{equation}
    {\ape}:= p_{e} \ominus p_{g}.
\end{equation}

In the proposed benchmark we provide scripts for automatic evaluation of both metrics and weight
their interest depending on the performed action.
For many common everyday objects
like bowls the rotation around their z axis is irrelevant because their symmetry is not distinguishable, even for humans . In this case we skip the angular component in the error calculation.
This metric is used for \textit{motion} sequences.
Additionally, it could be applied on other sequences as well, but this is not primarily targeted by this benchmark.

\subsubsection{Trajectory Error}

The other metric that we propose is based on the relative trajectory error between
the robot's end-effector and the interacting object over the period ($1:m$) of imitation. 
This results in a similar metric as proposed in \cite{sturm2012benchmark} for 
visual odometry using the root mean square error ($RMSE$): 

\begin{equation}
\label{eq:rmse}
    RMSE(\rpe_{1:m}) := \sqrt{\frac{1}{m} \sum^{m}_{j=1} \Vert \rpe_j \Vert^{2}}.
\end{equation}


\label{sec:baseline}
\begin{figure*}[h] \centering$
    \vspace{0.06in}
    \begin{array}{ccc}
        \includegraphics[height=.2\linewidth]{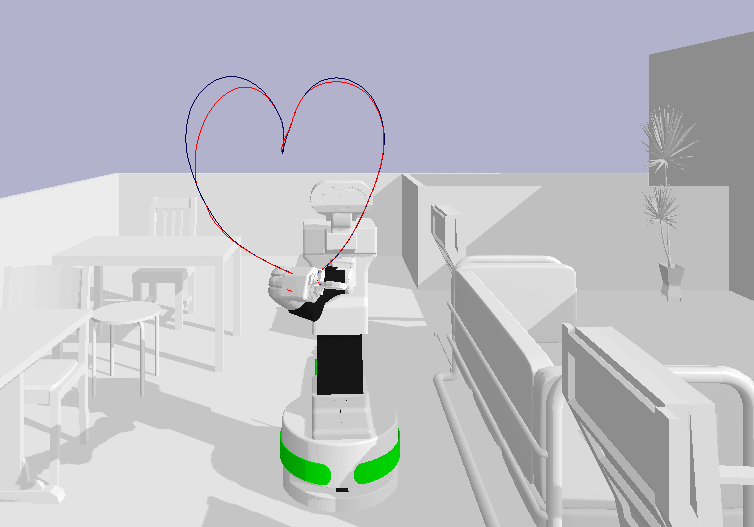} &
        \includegraphics[height=.2\linewidth]{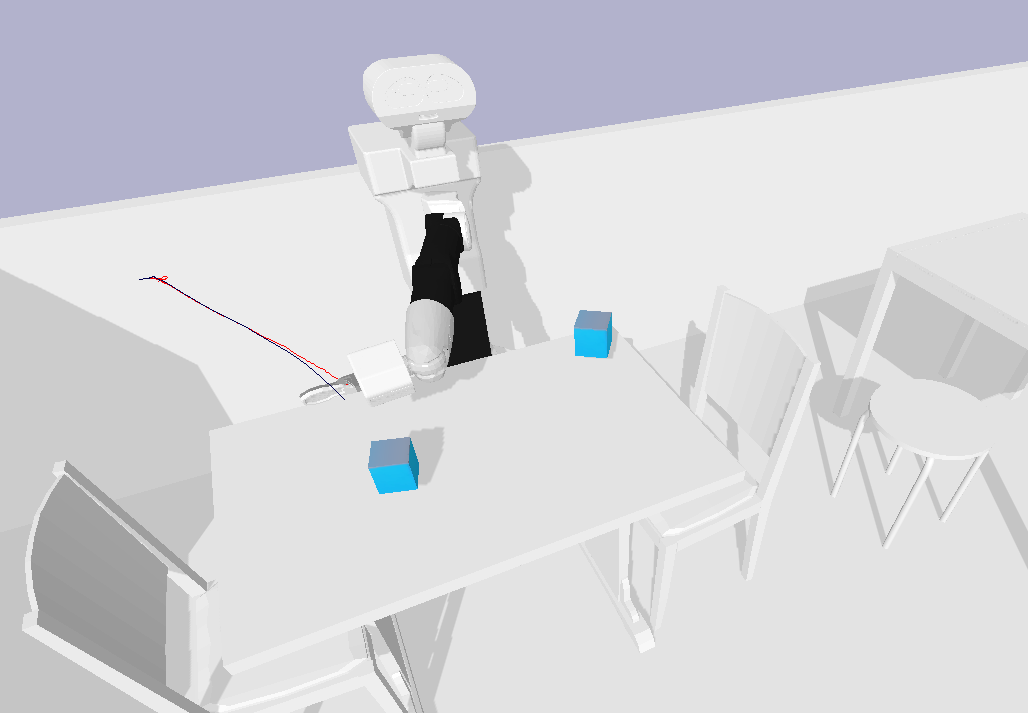} &
        \includegraphics[height=.2\linewidth]{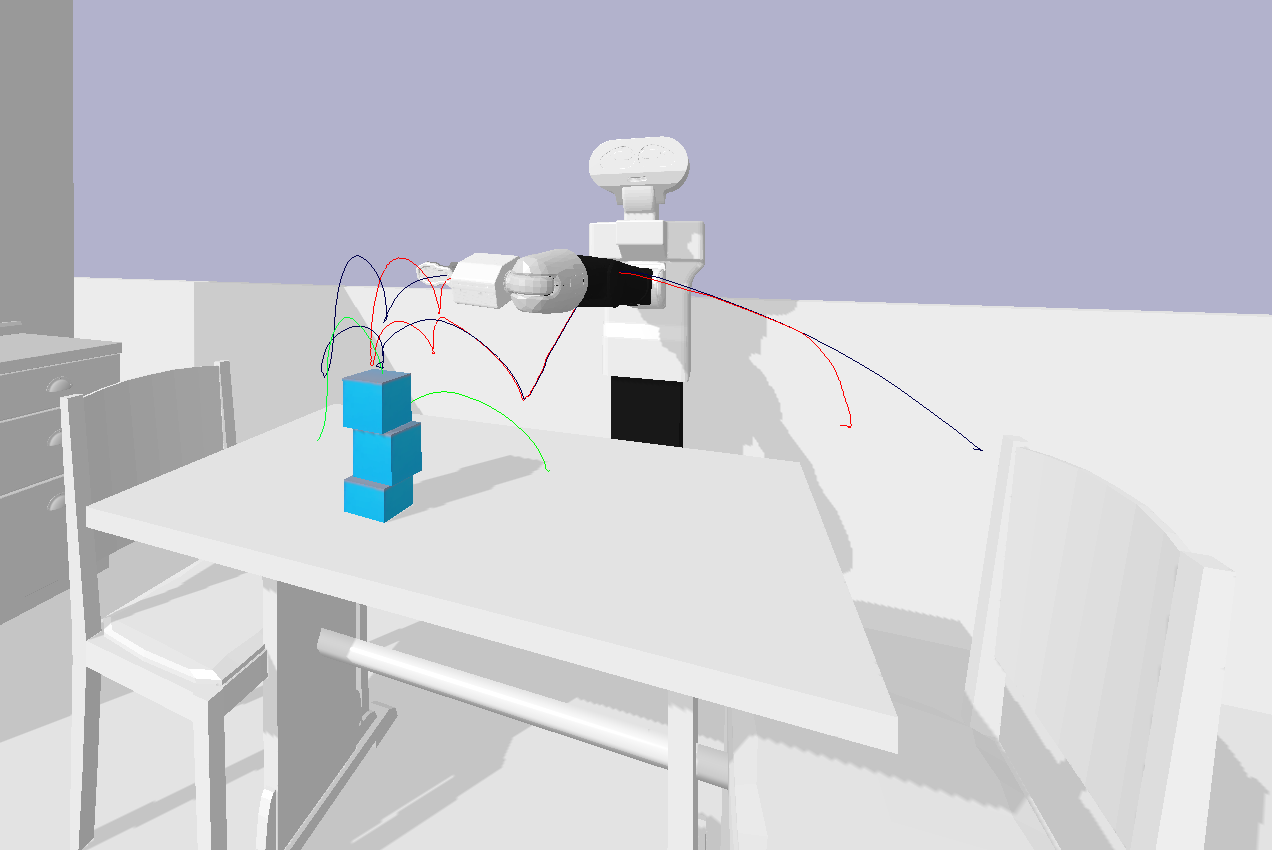} \\
        (a) & (b) & (c) 
    \end{array}
    $
    \caption{Trajectory comparison in simulation. The red line denotes the end-effector positions while the
  black line shows the ground truth positions of the demonstrator with a basic motions heart sequence $(a)$, a motion pick sequence $(b)$ and a sequential stack sequence $(c)$.}
    \label{fig:simulation}
\end{figure*}

To proof the validity of the proposed trajectory metrics and the benchmarking model, we 
implemented a simple approach for imitating human motions based on visual
observation. Such a scenario is visualized in \figname \ref{fig:simulation} $(a)$. For showing
the validity of the effect metric we took exemplary sequences and compared them against 
other demonstrated sequences involving the same set of objects.




\begin{table}[]
    \vspace{0.06in}
    \centering
    \caption{Evaluation of the absolute translation error (units are in $m$)}
    \newcommand{\bs}[1]{\textbf{#1}}
    \definecolor{SeqColor}{rgb}{0.9,0.9,0.9}
    \definecolor{White}{rgb}{1,1,1}
    \begin{tabular}{%
            >{\quad}
            p{0.225\linewidth}
            >{\RaggedLeft\arraybackslash}p{0.125\linewidth}
            >{\RaggedLeft\arraybackslash}p{0.125\linewidth}
            >{\RaggedLeft\arraybackslash}p{0.125\linewidth}
            >{\RaggedLeft\arraybackslash}p{0.125\linewidth}
        }
        \toprule
        \rowcolor{White}
        & \textbf{Min} & \textbf{Max} & \textbf{Mean}& \textbf{RMSE} \\
        \midrule
        \rowcolor{SeqColor}
        \hspace{-1.5em} Circle& & & &\\
        TIAGo KP        & {0.006}         & {0.673} & {0.105} &{0.160} \\
        TIAGo GT        & {0.007}         & {0.108} & {0.034} &{0.038} \\
        Sawyer KP          & {0.011}         & {0.755} & {0.110} &{0.174} \\
        Sawyer GT          & {0.003}         & {0.333} & {0.030} &{0.041} \\
         \rowcolor{SeqColor}
        \hspace{-1.5em} Rectangle& & & &\\
        TIAGo KP        & {0.010}         & {0.548} & {0.065} &{0.086} \\
        TIAGo GT        & {0.005}         & {0.139} & {0.028} &{0.032} \\
        Sawyer KP        & {0.009}         & {0.769}   & {0.061}   &{0.086} \\
        Sawyer GT        & {0.005}         & {0.386}   & {0.026}   &{0.033} \\
         \rowcolor{SeqColor}
        \hspace{-1.5em} Triangle& & & &\\
        TIAGo KP        & {0.015}            & {0.382} & {0.068} &{0.094} \\
        TIAGo GT        & {0.007}            & {0.112} & {0.024} &{0.034} \\
        Sawyer KP          & {0.014}         & {0.400} & {0.078} &{0.106} \\
        Sawyer GT          & {0.007}         & {0.114} & {0.025} &{0.036} \\
         \rowcolor{SeqColor}
        \hspace{-1.5em} Heart& & & &\\
        TIAGo KP       & {0.011}         & {0.362} & {0.054} &{0.073} \\
        TIAGo GT       & {0.007}         & {0.083} & {0.027} &{0.033} \\
        Sawyer KP         & {0.010}         & {0.701} & {0.057} &{0.085} \\
        Sawyer GT         & {0.005}         & {0.184} & {0.030} &{0.037} \\
         \rowcolor{SeqColor}
        \hspace{-1.5em} ZickZack& & & &\\
        TIAGo KP       & {0.024}         & {0.213} & {0.072} &{0.081} \\
        TIAGo GT       & {0.001}         & {0.098} & {0.036} &{0.043} \\
        Sawyer KP         & {0.022}         & {0.214} & {0.070} &{0.079} \\
        Sawyer GT         & {0.002}         & {0.108} & {0.035} &{0.043} \\
        \bottomrule
    \end{tabular}
    \label{tab:eval_ate}
\end{table}

For the basic motion sequences we evaluated the absolute trajectory error of the imitation. We use a keypoint detector for human pose estimation \cite{wei2016cpm} to estimate the hand positions in every frame 
of the sequence. The position of the right hand is projected in 3D space by using the depth 
channel of the corresponding pixel. The \ape\, of the first sequence of each set are with two robots shown in \tabname \ref{tab:eval_ate}. This table shows that the imitated hand poses with the robot's end-effector are reasonably accurate but subject for further improvement. We show the applied metric for the approached estimated hand keypoints (KP) and also in contrast what could be potentially be reachable with the proposed same initial setup by the robot with the groundtruth hand position (GT). The keypoint results are heavily influenced by outliers that occurred through projection errors of the corresponding 2D estimation to the corresponding depth value i.e. in cases where no depth could be estimated.



We also verified the validity of the effect evaluation using the \rpe for the imitation of a place sequence.
The robots are placed in front of the table in a similar position as the RGB-D camera was placed. The goal is to replicate the final state of the scene. For simplicity we attach the moved object to the end-effector position and computed the inverse kinematics to the goal location in order to compute the \rpe. For the TIAGo robot we got an average distance error of $0.047\,m$ and a rotational error of $0.013\,rad$ for the active object.  
The source code to reproduce the results is provided on the project page.

\section{Conclusion}
\label{sec:conclusion}

We proposed a novel benchmark for imitation learning tasks. A dataset recorded
with a RGB-D camera calibrated against a motion capturing system is coupled with
a simulated representation of the environment. Metrics for evaluation are proposed.
The goals of this benchmark are to foster comparability, reproducibility and the 
development of approaches for imitation learning tasks with a slight focus on visual imitation learning
approaches. The dataset does not just contain toy examples (like reaching or moving objects), 
but also more complex challenges to solve, for example ironing cloths and sequences for 
imitation on a trajectory level without object interactions. 
Simitate aims at keeping the entrance barrier low by providing a complete suite
with datasets, pre-trained models, integration into widely spread simulations and
simple visual baseline approaches as a starting point.
It can be extended by adding new tasks using an accessible testbed.
The effect metric will come to a limit on imitation learning tasks with 
soft-bodies like bed sheets or more liquids whereas for the trajectory metric 
one could argue that the same effects will be reached with the same motions, when not influenced
externally.


\subsubsection*{Acknowledgement}


We want to thank the Institute for Systems and Robotics at the Instituto 
Superior T\'{e}cnico, U. Lisboa, Portugal for enabling us to use the certified testbed
and supporting us in the use of the motion capturing system. 

\bibliographystyle{IEEEtranM}

\bibliography{main}

\ifx\mcitethebibliography\mciteundefinedmacro
\PackageError{IEEEtranM.bst}{mciteplus.sty has not been loaded}
{This bibstyle requires the use of the mciteplus package.}\fi
\begin{mcitethebibliography}{10}
\providecommand{\url}[1]{#1}
\csname url@samestyle\endcsname
\providecommand{\newblock}{\relax}
\providecommand{\bibinfo}[2]{#2}
\providecommand{\BIBentrySTDinterwordspacing}{\spaceskip=0pt\relax}
\providecommand{\BIBentryALTinterwordstretchfactor}{4}
\providecommand{\BIBentryALTinterwordspacing}{\spaceskip=\fontdimen2\font plus
\BIBentryALTinterwordstretchfactor\fontdimen3\font minus
  \fontdimen4\font\relax}
\providecommand{\BIBforeignlanguage}[2]{{%
\expandafter\ifx\csname l@#1\endcsname\relax
\typeout{** WARNING: IEEEtranM.bst: No hyphenation pattern has been}%
\typeout{** loaded for the language `#1'. Using the pattern for}%
\typeout{** the default language instead.}%
\else
\language=\csname l@#1\endcsname
\fi
#2}}
\providecommand{\BIBdecl}{\relax}
\BIBdecl

\bibitem{he2016deep}
K.~He, X.~Zhang, S.~Ren, and J.~Sun, ``Deep residual learning for image
  recognition,'' in \emph{Proceedings of the IEEE conference on computer vision
  and pattern recognition}, 2016, pp. 770--778\relax
\mciteBstWouldAddEndPuncttrue
\mciteSetBstMidEndSepPunct{;\space}{.}{\par\relax}\relax
\EndOfBibitem
\bibitem{krizhevsky2012imagenet}
A.~Krizhevsky, I.~Sutskever, and G.~E. Hinton, ``Imagenet classification with
  deep convolutional neural networks,'' in \emph{Advances in neural information
  processing systems}, 2012, pp. 1097--1105\relax
\mciteBstWouldAddEndPuncttrue
\mciteSetBstMidEndSepPunct{;\space}{.}{\par\relax}\relax
\EndOfBibitem
\bibitem{szegedy2015going}
C.~Szegedy, W.~Liu, Y.~Jia, P.~Sermanet, S.~Reed, D.~Anguelov, D.~Erhan,
  V.~Vanhoucke, and A.~Rabinovich, ``Going deeper with convolutions,'' in
  \emph{Proceedings of the IEEE conference on computer vision and pattern
  recognition}, 2015, pp. 1--9\relax
\mciteBstWouldAddEndPuncttrue
\mciteSetBstMidEndSepPunct{;\space}{.}{\par\relax}\relax
\EndOfBibitem
\bibitem{redmon2016you}
J.~Redmon, S.~Divvala, R.~Girshick, and A.~Farhadi, ``You only look once:
  Unified, real-time object detection,'' in \emph{Proceedings of the IEEE
  conference on computer vision and pattern recognition}, 2016, pp.
  779--788\relax
\mciteBstWouldAddEndPuncttrue
\mciteSetBstMidEndSepPunct{;\space}{.}{\par\relax}\relax
\EndOfBibitem
\bibitem{liu2016ssd}
W.~Liu, D.~Anguelov, D.~Erhan, C.~Szegedy, S.~Reed, C.-Y. Fu, and A.~C. Berg,
  ``Ssd: Single shot multibox detector,'' in \emph{European conference on
  computer vision}.\hskip 1em plus 0.5em minus 0.4em\relax Springer, 2016, pp.
  21--37\relax
\mciteBstWouldAddEndPuncttrue
\mciteSetBstMidEndSepPunct{;\space}{.}{\par\relax}\relax
\EndOfBibitem
\bibitem{he2017mask}
K.~He, G.~Gkioxari, P.~Doll{\'a}r, and R.~Girshick, ``Mask r-cnn,'' in
  \emph{Computer Vision (ICCV), 2017 IEEE International Conference on}.\hskip
  1em plus 0.5em minus 0.4em\relax IEEE, 2017, pp. 2980--2988\relax
\mciteBstWouldAddEndPuncttrue
\mciteSetBstMidEndSepPunct{;\space}{.}{\par\relax}\relax
\EndOfBibitem
\bibitem{badrinarayanan2015segnet}
V.~Badrinarayanan, A.~Kendall, and R.~Cipolla, ``Segnet: A deep convolutional
  encoder-decoder architecture for image segmentation,'' \emph{arXiv preprint
  arXiv:1511.00561}, 2015\relax
\mciteBstWouldAddEndPuncttrue
\mciteSetBstMidEndSepPunct{;\space}{.}{\par\relax}\relax
\EndOfBibitem
\bibitem{DeepMask}
P.~O. Pinheiro, R.~Collobert, and P.~Dollár, ``Learning to segment object
  candidates,'' in \emph{NIPS}, 2015\relax
\mciteBstWouldAddEndPuncttrue
\mciteSetBstMidEndSepPunct{;\space}{.}{\par\relax}\relax
\EndOfBibitem
\bibitem{cao2017realtime}
Z.~Cao, T.~Simon, S.-E. Wei, and Y.~Sheikh, ``Realtime multi-person 2d pose
  estimation using part affinity fields,'' in \emph{CVPR}, 2017\relax
\mciteBstWouldAddEndPuncttrue
\mciteSetBstMidEndSepPunct{;\space}{.}{\par\relax}\relax
\EndOfBibitem
\bibitem{simon2017hand}
T.~Simon, H.~Joo, I.~Matthews, and Y.~Sheikh, ``Hand keypoint detection in
  single images using multiview bootstrapping,'' in \emph{CVPR}, 2017\relax
\mciteBstWouldAddEndPuncttrue
\mciteSetBstMidEndSepPunct{;\space}{.}{\par\relax}\relax
\EndOfBibitem
\bibitem{wei2016cpm}
S.-E. Wei, V.~Ramakrishna, T.~Kanade, and Y.~Sheikh, ``Convolutional pose
  machines,'' in \emph{CVPR}, 2016\relax
\mciteBstWouldAddEndPuncttrue
\mciteSetBstMidEndSepPunct{;\space}{.}{\par\relax}\relax
\EndOfBibitem
\bibitem{lin2014microsoft}
T.-Y. Lin, M.~Maire, S.~Belongie, J.~Hays, P.~Perona, D.~Ramanan,
  P.~Doll{\'a}r, and C.~L. Zitnick, ``Microsoft coco: Common objects in
  context,'' in \emph{Computer Vision--ECCV 2014}.\hskip 1em plus 0.5em minus
  0.4em\relax Springer, 2014, pp. 740--755\relax
\mciteBstWouldAddEndPuncttrue
\mciteSetBstMidEndSepPunct{;\space}{.}{\par\relax}\relax
\EndOfBibitem
\bibitem{bernardin2008evaluating}
K.~Bernardin and R.~Stiefelhagen, ``Evaluating multiple object tracking
  performance: the clear mot metrics,'' \emph{Journal on Image and Video
  Processing}, vol. 2008, p.~1, 2008\relax
\mciteBstWouldAddEndPuncttrue
\mciteSetBstMidEndSepPunct{;\space}{.}{\par\relax}\relax
\EndOfBibitem
\bibitem{geiger2012we}
A.~Geiger, P.~Lenz, and R.~Urtasun, ``Are we ready for autonomous driving? the
  kitti vision benchmark suite,'' in \emph{Computer Vision and Pattern
  Recognition (CVPR), 2012 IEEE Conference on}.\hskip 1em plus 0.5em minus
  0.4em\relax IEEE, 2012, pp. 3354--3361\relax
\mciteBstWouldAddEndPuncttrue
\mciteSetBstMidEndSepPunct{;\space}{.}{\par\relax}\relax
\EndOfBibitem
\bibitem{sturm2012benchmark}
J.~Sturm, N.~Engelhard, F.~Endres, W.~Burgard, and D.~Cremers, ``A benchmark
  for the evaluation of rgb-d slam systems,'' in \emph{Intelligent Robots and
  Systems (IROS), 2012 IEEE/RSJ International Conference on}.\hskip 1em plus
  0.5em minus 0.4em\relax IEEE, 2012, pp. 573--580\relax
\mciteBstWouldAddEndPuncttrue
\mciteSetBstMidEndSepPunct{;\space}{.}{\par\relax}\relax
\EndOfBibitem
\bibitem{argall2009survey}
B.~D. Argall, S.~Chernova, M.~Veloso, and B.~Browning, ``A survey of robot
  learning from demonstration,'' \emph{Robotics and autonomous systems},
  vol.~57, no.~5, pp. 469--483, 2009\relax
\mciteBstWouldAddEndPuncttrue
\mciteSetBstMidEndSepPunct{;\space}{.}{\par\relax}\relax
\EndOfBibitem
\bibitem{osa2018algorithmic}
T.~Osa, J.~Pajarinen, G.~Neumann, J.~A. Bagnell, P.~Abbeel, J.~Peters
  \emph{et~al.}, ``An algorithmic perspective on imitation learning,''
  \emph{Foundations and Trends{\textregistered} in Robotics}, vol.~7, no. 1-2,
  pp. 1--179, 2018\relax
\mciteBstWouldAddEndPuncttrue
\mciteSetBstMidEndSepPunct{;\space}{.}{\par\relax}\relax
\EndOfBibitem
\bibitem{duan2017one}
Y.~Duan, M.~Andrychowicz, B.~Stadie, O.~J. Ho, J.~Schneider, I.~Sutskever,
  P.~Abbeel, and W.~Zaremba, ``One-shot imitation learning,'' in \emph{Advances
  in neural information processing systems}, 2017, pp. 1087--1098\relax
\mciteBstWouldAddEndPuncttrue
\mciteSetBstMidEndSepPunct{;\space}{.}{\par\relax}\relax
\EndOfBibitem
\bibitem{gangwani2018policy}
T.~Gangwani and J.~Peng, ``Policy optimization by genetic distillation,''
  2018\relax
\mciteBstWouldAddEndPuncttrue
\mciteSetBstMidEndSepPunct{;\space}{.}{\par\relax}\relax
\EndOfBibitem
\bibitem{gail2015towards}
T.~Gail, R.~Hoffmann, M.~Miezal, G.~Bleser, and S.~Leyendecker, ``Towards
  bridging the gap between motion capturing and biomechanical optimal control
  simulations,'' in \emph{Thematic Conference on Multibody Dynamics},
  2015\relax
\mciteBstWouldAddEndPuncttrue
\mciteSetBstMidEndSepPunct{;\space}{.}{\par\relax}\relax
\EndOfBibitem
\bibitem{Ross2010}
\BIBentryALTinterwordspacing\relax S.~Ross, G.~J. Gordon, and J.~A. Bagnell,
  ``{A Reduction of Imitation Learning and Structured Prediction to No-Regret
  Online Learning},'' \emph{Proceedings of AISTATS}, vol.~15, pp. 627--635,
  2010. [Online]. Available: \url{http://arxiv.org/abs/1011.0686}\relax
\mciteBstWouldAddEndPunctfalse
\mciteSetBstMidEndSepPunct{~;\space}{}{\par\BIBentrySTDinterwordspacing}\relax
\EndOfBibitem
\bibitem{Laskey2017}
\BIBentryALTinterwordspacing\relax M.~Laskey, J.~Lee, R.~Fox, A.~Dragan, and
  K.~Goldberg, ``{DART: Noise Injection for Robust Imitation Learning},'' no.
  CoRL, pp. 1--14, 2017. [Online]. Available:
  \url{http://arxiv.org/abs/1703.09327}\relax
\mciteBstWouldAddEndPunctfalse
\mciteSetBstMidEndSepPunct{~;\space}{}{\par\BIBentrySTDinterwordspacing}\relax
\EndOfBibitem
\bibitem{ho2016generative}
J.~Ho and S.~Ermon, ``Generative adversarial imitation learning,'' in
  \emph{Advances in Neural Information Processing Systems}, 2016, pp.
  4565--4573\relax
\mciteBstWouldAddEndPuncttrue
\mciteSetBstMidEndSepPunct{;\space}{.}{\par\relax}\relax
\EndOfBibitem
\bibitem{DBLP:conf/rss/YuFDXZAL18}
\BIBentryALTinterwordspacing\relax T.~Yu, C.~Finn, S.~Dasari, A.~Xie, T.~Zhang,
  P.~Abbeel, and S.~Levine, ``One-shot imitation from observing humans via
  domain-adaptive meta-learning,'' in \emph{Robotics: Science and Systems XIV,
  Carnegie Mellon University, Pittsburgh, Pennsylvania, USA, June 26-30, 2018},
  H.~Kress{-}Gazit, S.~Srinivasa, T.~Howard, and N.~Atanasov, Eds., 2018.
  [Online]. Available:
  \url{http://www.roboticsproceedings.org/rss14/p02.html}\relax
\mciteBstWouldAddEndPunctfalse
\mciteSetBstMidEndSepPunct{~;\space}{}{\par\BIBentrySTDinterwordspacing}\relax
\EndOfBibitem
\bibitem{DBLP:journals/corr/abs-1810-05017}
\BIBentryALTinterwordspacing\relax T.~L. Paine, S.~G. Colmenarejo, Z.~Wang,
  S.~E. Reed, Y.~Aytar, T.~Pfaff, M.~W. Hoffman, G.~Barth{-}Maron, S.~Cabi,
  D.~Budden, and N.~de~Freitas, ``One-shot high-fidelity imitation: Training
  large-scale deep nets with {RL},'' \emph{CoRR}, vol. abs/1810.05017, 2018.
  [Online]. Available: \url{http://arxiv.org/abs/1810.05017}\relax
\mciteBstWouldAddEndPunctfalse
\mciteSetBstMidEndSepPunct{~;\space}{}{\par\BIBentrySTDinterwordspacing}\relax
\EndOfBibitem
\bibitem{DBLP:conf/nips/DuanASHSSAZ17}
\BIBentryALTinterwordspacing\relax Y.~Duan, M.~Andrychowicz, B.~C. Stadie,
  J.~Ho, J.~Schneider, I.~Sutskever, P.~Abbeel, and W.~Zaremba, ``One-shot
  imitation learning,'' in \emph{Advances in Neural Information Processing
  Systems 30: Annual Conference on Neural Information Processing Systems 2017,
  4-9 December 2017, Long Beach, CA, {USA}}, I.~Guyon, U.~von Luxburg,
  S.~Bengio, H.~M. Wallach, R.~Fergus, S.~V.~N. Vishwanathan, and R.~Garnett,
  Eds., 2017, pp. 1087--1098. [Online]. Available:
  \url{http://papers.nips.cc/paper/6709-one-shot-imitation-learning}\relax
\mciteBstWouldAddEndPunctfalse
\mciteSetBstMidEndSepPunct{~;\space}{}{\par\BIBentrySTDinterwordspacing}\relax
\EndOfBibitem
\bibitem{bates2017line}
T.~Bates, K.~Ramirez-Amaro, T.~Inamura, and G.~Cheng, ``On-line simultaneous
  learning and recognition of everyday activities from virtual reality
  performances,'' in \emph{Intelligent Robots and Systems (IROS), 2017 IEEE/RSJ
  International Conference on}.\hskip 1em plus 0.5em minus 0.4em\relax IEEE,
  2017, pp. 3510--3515\relax
\mciteBstWouldAddEndPuncttrue
\mciteSetBstMidEndSepPunct{;\space}{.}{\par\relax}\relax
\EndOfBibitem
\bibitem{amaro2014bootstrapping}
K.~R. Amaro, T.~Inamura, E.~Dean-Le{\'o}n, M.~Beetz, and G.~Cheng,
  ``Bootstrapping humanoid robot skills by extracting semantic representations
  of human-like activities from virtual reality,'' in \emph{IEEE-RAS
  International Conference on Humanoid Robots}, 2014\relax
\mciteBstWouldAddEndPuncttrue
\mciteSetBstMidEndSepPunct{;\space}{.}{\par\relax}\relax
\EndOfBibitem
\bibitem{mizuchi2017cloud}
Y.~Mizuchi and T.~Inamura, ``Cloud-based multimodal human-robot interaction
  simulator utilizing ros and unity frameworks,'' in \emph{System Integration
  (SII), 2017 IEEE/SICE International Symposium on}.\hskip 1em plus 0.5em minus
  0.4em\relax IEEE, 2017, pp. 948--955\relax
\mciteBstWouldAddEndPuncttrue
\mciteSetBstMidEndSepPunct{;\space}{.}{\par\relax}\relax
\EndOfBibitem
\bibitem{weinzaepfel2016human}
P.~Weinzaepfel, X.~Martin, and C.~Schmid, ``Human action localization with
  sparse spatial supervision,'' \emph{arXiv preprint arXiv:1605.05197},
  2016\relax
\mciteBstWouldAddEndPuncttrue
\mciteSetBstMidEndSepPunct{;\space}{.}{\par\relax}\relax
\EndOfBibitem
\bibitem{pirsiavash2012detecting}
H.~Pirsiavash and D.~Ramanan, ``Detecting activities of daily living in
  first-person camera views,'' in \emph{Computer Vision and Pattern Recognition
  (CVPR), 2012 IEEE Conference on}.\hskip 1em plus 0.5em minus 0.4em\relax
  IEEE, 2012, pp. 2847--2854\relax
\mciteBstWouldAddEndPuncttrue
\mciteSetBstMidEndSepPunct{;\space}{.}{\par\relax}\relax
\EndOfBibitem
\bibitem{zhang2016rgb}
J.~Zhang, W.~Li, P.~O. Ogunbona, P.~Wang, and C.~Tang, ``Rgb-d-based action
  recognition datasets: A survey,'' \emph{Pattern Recognition}, vol.~60, pp.
  86--105, 2016\relax
\mciteBstWouldAddEndPuncttrue
\mciteSetBstMidEndSepPunct{;\space}{.}{\par\relax}\relax
\EndOfBibitem
\bibitem{Codevilla2018}
F.~Codevilla, M.~M{\"u}ller, A.~L{\'o}pez, V.~Koltun, and A.~Dosovitskiy,
  ``End-to-end driving via conditional imitation learning,'' in
  \emph{International Conference on Robotics and Automation (ICRA)}, 2018\relax
\mciteBstWouldAddEndPuncttrue
\mciteSetBstMidEndSepPunct{;\space}{.}{\par\relax}\relax
\EndOfBibitem
\bibitem{zhang2016query}
J.~Zhang and K.~Cho, ``Query-efficient imitation learning for end-to-end
  autonomous driving,'' \emph{arXiv preprint arXiv:1605.06450}, 2016\relax
\mciteBstWouldAddEndPuncttrue
\mciteSetBstMidEndSepPunct{;\space}{.}{\par\relax}\relax
\EndOfBibitem
\bibitem{gupta2015visual}
S.~Gupta and J.~Malik, ``Visual semantic role labeling,'' \emph{arXiv preprint
  arXiv:1505.04474}, 2015\relax
\mciteBstWouldAddEndPuncttrue
\mciteSetBstMidEndSepPunct{;\space}{.}{\par\relax}\relax
\EndOfBibitem
\bibitem{koppula2013learning}
H.~S. Koppula, R.~Gupta, and A.~Saxena, ``Learning human activities and object
  affordances from rgb-d videos,'' \emph{The International Journal of Robotics
  Research}, vol.~32, no.~8, pp. 951--970, 2013\relax
\mciteBstWouldAddEndPuncttrue
\mciteSetBstMidEndSepPunct{;\space}{.}{\par\relax}\relax
\EndOfBibitem
\bibitem{wu2013online}
Y.~Wu, J.~Lim, and M.-H. Yang, ``Online object tracking: A benchmark,'' in
  \emph{Proceedings of the IEEE conference on computer vision and pattern
  recognition}, 2013, pp. 2411--2418\relax
\mciteBstWouldAddEndPuncttrue
\mciteSetBstMidEndSepPunct{;\space}{.}{\par\relax}\relax
\EndOfBibitem
\bibitem{milan2016mot16}
A.~Milan, L.~Leal-Taix{\'e}, I.~Reid, S.~Roth, and K.~Schindler, ``Mot16: A
  benchmark for multi-object tracking,'' \emph{arXiv preprint
  arXiv:1603.00831}, 2016\relax
\mciteBstWouldAddEndPuncttrue
\mciteSetBstMidEndSepPunct{;\space}{.}{\par\relax}\relax
\EndOfBibitem
\bibitem{leitner2017acrv}
J.~Leitner, A.~W. Tow, N.~S{\"u}nderhauf, J.~E. Dean, J.~W. Durham, M.~Cooper,
  M.~Eich, C.~Lehnert, R.~Mangels, C.~McCool \emph{et~al.}, ``The acrv picking
  benchmark: A robotic shelf picking benchmark to foster reproducible
  research,'' in \emph{Robotics and Automation (ICRA), 2017 IEEE International
  Conference on}.\hskip 1em plus 0.5em minus 0.4em\relax IEEE, 2017, pp.
  4705--4712\relax
\mciteBstWouldAddEndPuncttrue
\mciteSetBstMidEndSepPunct{;\space}{.}{\par\relax}\relax
\EndOfBibitem
\bibitem{zhang2017deep}
T.~Zhang, Z.~McCarthy, O.~Jow, D.~Lee, K.~Goldberg, and P.~Abbeel, ``Deep
  imitation learning for complex manipulation tasks from virtual reality
  teleoperation,'' \emph{arXiv preprint arXiv:1710.04615}, 2017\relax
\mciteBstWouldAddEndPuncttrue
\mciteSetBstMidEndSepPunct{;\space}{.}{\par\relax}\relax
\EndOfBibitem
\bibitem{Villani2018}
V.~Villani, B.~Capelli, and L.~Sabattini, ``{Use of Virtual Reality for the
  Evaluation of Human-Robot Interaction Systems in Complex Scenarios},'' vol.
  Proceeding, pp. 422--427, 2018\relax
\mciteBstWouldAddEndPuncttrue
\mciteSetBstMidEndSepPunct{;\space}{.}{\par\relax}\relax
\EndOfBibitem
\bibitem{wisspeintner2009robocup}
T.~Wisspeintner, T.~Van Der~Zant, L.~Iocchi, and S.~Schiffer, ``Robocup@ home:
  Scientific competition and benchmarking for domestic service robots,''
  \emph{Interaction Studies}, vol.~10, no.~3, pp. 392--426, 2009\relax
\mciteBstWouldAddEndPuncttrue
\mciteSetBstMidEndSepPunct{;\space}{.}{\par\relax}\relax
\EndOfBibitem
\bibitem{lima2016rockin}
P.~U. Lima, D.~Nardi, G.~K. Kraetzschmar, R.~Bischoff, and M.~Matteucci,
  ``Rockin and the european robotics league: building on robocup best practices
  to promote robot competitions in europe,'' in \emph{Robot World Cup}.\hskip
  1em plus 0.5em minus 0.4em\relax Springer, 2016, pp. 181--192\relax
\mciteBstWouldAddEndPuncttrue
\mciteSetBstMidEndSepPunct{;\space}{.}{\par\relax}\relax
\EndOfBibitem
\bibitem{Alissandrakis2007}
A.~Alissandrakis, C.~L. Nehaniv, and K.~Dautenhahn, ``{Correspondence mapping
  induced state and action metrics for robotic imitation},'' \emph{IEEE
  Transactions on Systems, Man, and Cybernetics, Part B: Cybernetics}, vol.~37,
  no.~2, pp. 299--307, 2007\relax
\mciteBstWouldAddEndPuncttrue
\mciteSetBstMidEndSepPunct{;\space}{.}{\par\relax}\relax
\EndOfBibitem
\bibitem{Alissandrakis2006}
\BIBentryALTinterwordspacing\relax A.~Alissandrakis, C.~L. Nehaniv,
  K.~Dautenhahn, and J.~Saunders, ``{Evaluation of robot imitation attempts:
  comparison of the system's and the human's perspectives},'' \emph{Proceedings
  of the 1st ACM SIGCHI/SIGART conference on Human-robot interaction}, pp.
  134--141, 2006. [Online]. Available:
  \url{http://portal.acm.org/citation.cfm?id=1121265}\relax
\mciteBstWouldAddEndPunctfalse
\mciteSetBstMidEndSepPunct{~;\space}{}{\par\BIBentrySTDinterwordspacing}\relax
\EndOfBibitem
\bibitem{zhang2000flexible}
Z.~Zhang, ``A flexible new technique for camera calibration,'' \emph{IEEE
  Transactions on pattern analysis and machine intelligence}, vol.~22,
  2000\relax
\mciteBstWouldAddEndPuncttrue
\mciteSetBstMidEndSepPunct{;\space}{.}{\par\relax}\relax
\EndOfBibitem
\bibitem{Man+18}
K.-K. Maninis, S.~Caelles, J.~Pont-Tuset, and L.~{Van Gool}, ``Deep extreme
  cut: From extreme points to object segmentation,'' in \emph{Computer Vision
  and Pattern Recognition (CVPR)}, 2018\relax
\mciteBstWouldAddEndPuncttrue
\mciteSetBstMidEndSepPunct{;\space}{.}{\par\relax}\relax
\EndOfBibitem
\bibitem{koenig2004design}
N.~P. Koenig and A.~Howard, ``Design and use paradigms for gazebo, an
  open-source multi-robot simulator.'' in \emph{IROS}, vol.~4.\hskip 1em plus
  0.5em minus 0.4em\relax Citeseer, 2004, pp. 2149--2154\relax
\mciteBstWouldAddEndPuncttrue
\mciteSetBstMidEndSepPunct{;\space}{.}{\par\relax}\relax
\EndOfBibitem
\bibitem{coumans2019}
E.~Coumans and Y.~Bai, ``Pybullet, a python module for physics simulation for
  games, robotics and machine learning,'' \url{http://pybullet.org},
  2016--2019\relax
\mciteBstWouldAddEndPuncttrue
\mciteSetBstMidEndSepPunct{;\space}{.}{\par\relax}\relax
\EndOfBibitem
\bibitem{kummerle2009measuring}
R.~K{\"u}mmerle, B.~Steder, C.~Dornhege, M.~Ruhnke, G.~Grisetti, C.~Stachniss,
  and A.~Kleiner, ``On measuring the accuracy of slam algorithms,''
  \emph{Autonomous Robots}, vol.~27, no.~4, p. 387, 2009\relax
\mciteBstWouldAddEndPuncttrue
\mciteSetBstMidEndSepPunct{;\space}{.}{\par\relax}\relax
\EndOfBibitem
\end{mcitethebibliography}

\end{document}